\documentclass[letterpaper]{article} 
\usepackage{aaai25}  
\usepackage{times}  
\usepackage{helvet}  
\usepackage{courier}  
\usepackage[hyphens]{url}  
\usepackage{graphicx} 
\usepackage{natbib}  
\usepackage{caption} 
\usepackage{algorithm}
\usepackage{algorithmic}

\usepackage{newfloat}
\usepackage{listings}
\DeclareCaptionStyle{ruled}{labelfont=normalfont,labelsep=colon,strut=off} 
\floatstyle{ruled}
\newfloat{listing}{tb}{lst}{}
\floatname{listing}{Listing}
\title{\LTLf Adaptive Synthesis for Multi-Tier Goals in Nondeterministic Domains}
\author {
    Giuseppe De Giacomo\textsuperscript{\rm 1, \rm 2},
    Gianmarco Parretti\textsuperscript{\rm 2},
    Shufang Zhu\textsuperscript{\rm 3}
}
\affiliations {
    \textsuperscript{\rm 1}University of Oxford\\
    \textsuperscript{\rm 2}University of Rome ``La Sapienza''\\
    \textsuperscript{\rm 3}University of Liverpool\\
    giuseppe.degiacomo@cs.ox.ac.uk, parretti@diag,uniroma1.it, shufang.zhu@liverpool.ac.uk
}

\usepackage{bibentry}

\usepackage{xspace}

\newcommand{\myi}{(\emph{i})\xspace}
\newcommand{\myii}{(\emph{ii})\xspace}
\newcommand{\myiii}{(\emph{iii})\xspace}
\newcommand{\myiv}{(\emph{iv})\xspace}

\newcommand{\tst}{\text{ s.t. }}
\newcommand{\tiff}{\text{ iff }}
\newcommand{\LTLftitle}{\textbf{LTL$_f$}\xspace}
\newcommand{\LTLf}{{\sc ltl}$_f$\xspace}

\newcommand{\LTL}{{\sc ltl}\xspace}
\newcommand{\trace}{\pi} 
\newcommand{\pmodels}{\models_{\P}}

\newcommand{\last}{\mathtt{lst}}

\newcommand{\true}{\mathit{true}}

\newcommand{\false}{\mathit{false}}

\newcommand{\Next}{\raisebox{-0.27ex}{\LARGE$\circ$}}
\newcommand{\Wnext}{\raisebox{-0.27ex}{\LARGE$\bullet$}}
\newcommand{\Until}{\mathop{\U}}

\newcommand{\A}{\mathcal{A}}

\renewcommand{\P}{\mathcal{P}} 

\newcommand{\U}{\mathcal{U}}

\newcommand{\G}{\mathcal{G}}
\newcommand{\F}{\mathcal{F}}
\newcommand{\T}{\mathcal{T}}
\renewcommand{\L}{\mathcal{L}}
\renewcommand{\H}{\mathcal{H}}

\newcommand{\DFA}{{\sc dfa}\xspace}
\newcommand{\DFAs}{{\sc dfa}s\xspace}

\renewcommand{\part}{\mapsto}

\newcommand{\exptime}{{\sc exptime}\xspace}
\newcommand{\twoexptime}{2{\sc exptime}\xspace}

\newcommand{\Play}{\mathtt{Play}}
\newcommand\run{{\texttt{Run}}}

\newcommand{\w}{\mathrm{W}}
\renewcommand{\c}{\mathrm{C}}

\newcommand{\wc}{\mathrm{WP}}

\newcommand{\act}{Act}
\newcommand{\react}{React}

\newcommand{\win}{\textit{win}}
\newcommand{\pend}{\textit{pend}}
\newcommand{\lose}{\textit{lose}}

\newcommand{\val}[1]{\mathrm{val}_{#1}}

\newcommand{\agerr}{s^{ag}_{err}}
\newcommand{\enverr}{s^{env}_{err}}
\newcommand{\bigagerr}{AgErr}
\newcommand{\bigenverr}{EnvErr}

\newcommand{\FOND}{{\sc fond}\xspace}
\newcommand{\PDDL}{{\sc pddl}\xspace}

\usepackage{paralist}
\usepackage{amssymb}
\usepackage{amsmath}
\usepackage{amsthm}
\usepackage{enumitem}

\newtheorem{definition}{Definition}

\newtheorem{theorem}{Theorem}

\begin{document}

\maketitle

\begin{abstract}
We study a variant of \LTLf synthesis that synthesizes adaptive strategies for achieving a multi-tier goal, consisting of multiple increasingly challenging \LTLf objectives in nondeterministic planning domains. Adaptive strategies are strategies that at any point of their execution \myi enforce the satisfaction of as many objectives as possible in the multi-tier goal, and \myii exploit possible cooperation from the environment to satisfy as many as possible of the remaining ones. This happens dynamically: if the environment cooperates \myii and an objective becomes enforceable \myi, then our strategies will enforce it. We provide a game-theoretic technique to compute adaptive strategies that is sound and complete. Notably, our technique is polynomial, in fact quadratic, in the number of objectives. In other words, it handles multi-tier goals with only a minor overhead compared to standard \LTLf synthesis.
\end{abstract}

\section{Introduction~\label{sec:intro}}

There has been a growing interest in Reasoning about Actions, Planning, and Sequential Decision Making, to techniques from formal methods for strategic reasoning such as reactive synthesis \cite{PnueliR89,fijalkow2024}, which can play a crucial role in developing autonomous AI systems capable of self-programming their actions to complete desired goals. Such capabilities are particularly important when these systems operate within complex, dynamic environments, in particular, with high levels of nondeterminism. 

Most of the literature on strategy synthesis~\cite{fijalkow2024,DegVa15,DeGR18,CBM19} and planning~\cite{GhallabNauTraverso2016,Haslum2019} assumes that the AI system or the \emph{agent}, is working towards a singular goal (typically specified, e.g., in standard reachability \cite{Haslum2019}, or in Temporal Logics \cite{AminofDMR19,CBM19}). The focus of planners/synthesizers is generally on finding a plan/strategy as quickly as possible~\cite{Haslum2019,GeBo2013,fijalkow2024}. However, in real-world applications, users often struggle to specify an agent goal accurately using only high-level characteristics. Preferences-based planning~(PBP)~\cite{son2006planning,KR06BienvenuM,jorge2008planning} addresses this challenge by allowing users to specify preferences over the plans. The aim in PBP is to produce plans that satisfy as many user-specified preferences as possible, such as goal preferences, action preferences, state preferences and temporal preferences~\cite{bai-bac-mci-aij09}. 
Planning with soft goals~\cite{keyder2009soft} is a simple model of PBP, which aims to maximize utility in cases where the agent may not be able to achieve all its goals. PBP frameworks optimize the selection of objectives, deciding which to achieve and which to discard, with the discarded ones being those that do not contribute to optimal preference satisfaction. 

In this paper we focus on Fully Observable Nondeterministic Domains~(\FOND)~\cite{GeBo2013} and consider temporal preferences that are ordinal~\cite{son2006planning,KR06BienvenuM}. Specifically, the agent goals are multi-tier goals consisting of a multi-tier hierarchy of increasingly challenging objectives expressed in Linear Temporal Logic on finite traces~(\LTLf)~\cite{DegVa13}. This enables users to specify when a strategy is better than another by simply considering additional tasks,
%
e.g., a strategy that enforces ``clean room A and deliver package B, $\varphi_2 = \Diamond(clean_A) \land \Diamond(deliver_B)$'' is better than another strategy that only enforces ``clean room A $\varphi_1 = \Diamond(clean_A)$''. 
%
If we simply aim at satisfying as many objectives as possible, considering that they are in a hierarchy of increasingly challenging objectives, the problem reduces to finding the most challenging objective for which a winning strategy can be synthesized. This can be done with off-the-shelf techniques \cite{DeGR18}.

However, as observed in \cite{shaparau2006contingent} in the context of \FOND for reachability goals, in nondeterministic domains it is interesting to make the optimality criteria depend on the state of the domain and the response of the environment. Importantly, all objectives must remain under consideration at all instants, as objectives that are currently unachievable could become feasible later due to nondeterministic responses. For example while currently only 
$\varphi_1 = \Diamond(clean_A)$ is achievable for any environment response, a specific environment response might make $\varphi_2 = \Diamond(clean_A) \land \Diamond(deliver_B)$ achievable as well.

Embracing this intuition, 
we seek \emph{adaptive strategies}\footnote{We use the term \emph{adaptive} to qualify strategies with the three properties above, but the term ``adaptive" can be used to denote very different concepts, see  e.g., ~\cite{DBKMSU14,CiolekDPS20,Rodriguez024}.}~(plans) that at any point during execution \myi enforce the satisfaction of as many \LTLf objectives as possible, and \myii exploit possible environment cooperation to satisfy as many remaining ones as possible. This happens dynamically: if the environment cooperates \myii and an objective becomes enforceable \myi, then our strategies will enforce it.



Formally we base our approach on \emph{best-effort strategies}~\cite{ADLMR20,ADLMR21,ADR21}.
Best-effort strategies are strategies that enforce objective satisfaction whenever possible, and do nothing that needlessly prevent satisfying the objective, otherwise. Intuitively, if an objective that is not enforceable now but could still become enforceable later during execution due to possible cooperative environment response, best-effort strategies will exploit such response to enforce objective satisfaction. In this sense we can think
best-effort strategies as adaptive strategies for a single objective.
Note, if winning strategies exist, then best-effort strategies are exactly the winning strategies. However while winning strategies may not exist, best effort strategies always do. 
Best-effort strategies for \LTLf objectives can be computed in worst-case 2EXPTIME, just as for winning strategies~(best-effort synthesis is 2EXPTIME-complete, just as in standard synthesis)~\cite{ADR21}. Note that such worst-case blowup depends on the need to build a \DFA for the \LTLf formula, and occurs only rarely in practice, see, e.g.~\cite{DegVa15,CamachoIKVM19,ZGPV20,GeattiMR24}.

Leveraging results on best-effort strategies, we study synthesis of adaptive strategies for  multi-tier goals. Specifically we want to synthesize a strategy that in every history fulfills a maximal number of objectives. Hence, an adaptive strategy for a multi-tier goal keeps all objectives active by winning as many objectives as possible, regardless of environment responses, and not ruling out completion for as many as possible of the remaining ones. In particular, we prove that such strategies (as best-effort strategies for single objectives) always exist~(see  Theorem~\ref{thm:existence}).


Our main contribution is a game-theoretic technique to compute an adaptive strategy for multi-tier goals that is sound and complete, as well as practically efficient. Our technique first identifies the most challenging objective that the agent can win in spite of the environment, which is referred to as \emph{maximally winning objective}. This objective represents the highest tier, i.e., most challenging, that the agent can enforce against all possible environment behaviors, hence also enforcing lower-tier, i.e., easier, objectives in the multi-tier goal hierarchy. Then, we identify the \emph{maximally winning-pending objective}, which is higher than (or equal to) the maximally winning one. In this way we can win for the maximally winning objective and leave open the possibility to further win as many as possible of the higher-tier objectives if the environment cooperates. To compute the maximally winning-pending objective, our approach requires consideration of only the maximally winning objective and each of those more challenging objectives. As a result, our technique remains worst-case 2EXPTIME-complete in the size of the objectives in the \LTLf multi-tier goal~as standard \LTLf synthesis, but notably it is \emph{polynomial}, in fact quadratic, in the number of objectives~(Theorem~\ref{thm:complexity}). In this way, we are able to handle multi-tier goals with only a minor overhead compared to standard \LTLf synthesis.

\section{Preliminaries~\label{sec:preliminaries}}

A \emph{trace} over an alphabet of symbols $\Sigma$ is a finite or infinite sequence of elements from $\Sigma$. The empty trace is $\lambda$. Traces are indexed starting at zero, and we write $\trace = \trace_0 \trace_1 \cdots$.  The length of a trace is $|\trace|$. For a finite trace $\trace$, we denote by $\last(\trace)$ the index of the last element of $\trace$, i.e., $|\trace| - 1$. 


\paragraph{\LTLftitle.} Linear Temporal Logic on finite traces (\LTLf) is a specification language to express temporal properties for finite and non-empty traces~\cite{DegVa13}. \LTLf shares the syntax with \LTL~\cite{Pnu77}, but is interpreted over finite non-empty traces. Given a set of atoms $AP$, the \LTLf formulas over $AP$ are generated as follows: \\
{\centerline{$\varphi ::= p \mid \varphi_1 \wedge \varphi_2 \mid \neg \varphi \mid  
 \Next \varphi \mid \varphi_1 \Until \varphi_2.$}}
$p \in AP$ is an \textit{atom}, $\Next$~(\emph{Next}), and $\Until$~(\emph{Until}) are temporal operators. 
We use standard Boolean abbreviations such as $\vee$~(or) and $\supset$~(implies), $\true$ and $\false$. Moreover, we define the following abbreviations \emph{Weak Next} $\Wnext \varphi \equiv \neg \Next \neg \varphi$, \emph{Eventually} $\Diamond \varphi \equiv \true \Until \varphi$ and \emph{Always} $\Box \varphi \equiv \neg \Diamond \neg \varphi$.
The size of $\varphi$, written $|\varphi|$, is the number of all its subformulas. 

\LTLf formulas are interpreted over finite non-empty traces $\trace$ over the alphabet $\Sigma = 2^{AP}$, i.e., the alphabet consisting of the propositional interpretations of the atoms. Thus, for $i \leq \last({\trace})$, we have that $\trace_i \in 2^{AP}$ is the $i$-th interpretation of $\trace$. An \LTLf formula $\varphi$ \emph{holds} at instant $i$ of a trace $\pi$ is defined inductively on the structure of $\varphi$ as:
\begin{compactitem}
	\item 
	$\trace, i \models p \tiff p \in \trace_i$;
	\item 
	$\trace, i \models \lnot \varphi \tiff \trace, i \not\models \varphi\nonumber$;
	\item 
	$\trace, i \models \varphi_1 \wedge \varphi_2 \tiff \trace, i \models \varphi_1 \text{ and } \trace, i \models \varphi_2$;
	\item 
	$\pi, i \models \Next\varphi \tiff  i< \last(\pi)$ and $\trace,i+1 \models \varphi$;
	\item 
	$\pi, i \models \varphi_1 \Until \varphi_2$ iff $\exists j$ such that $i \leq j \leq \last(\pi)$ and $\pi,j \models\varphi_2$, and $\forall k, i\le k < j$ we have that $\pi, k \models \varphi_1$.
\end{compactitem}

We say that $\pi$ \emph{satisfies} $\varphi$, written as $\pi \models \varphi$, if $\pi, 0 \models \varphi$.

\paragraph{Nondeterministic Planning Domains.} Following~\cite{DPZ23}, a \emph{nondeterministic planning domain} is a tuple $\P = (2^\F, s_0, \act, \react, \alpha, \beta, \delta)$, where: $\F$ is a finite set of fluents, $|\F|$ is the size of $\P$, and $2^\F$ is the state space; $s_0 \in 2^\F $is the initial state; $\act$ is a finite set of agent actions; $\react$ is a finite set of environment reactions; $\alpha:2^\F \rightarrow 2^{\act}$ denotes agent action preconditions; $\beta: 2^\F \times \act \rightarrow 2^{\react}$ denotes environment reaction preconditions; and $\delta: 2^\F \times \act \times \react \mapsto 2^{\F}$ is the transition function such that $\delta(s,a,r)$ is defined iff $a \in \alpha(s)$ and $r \in \beta(s,a)$.
As in~\cite{DPZ23}, we require nondeterministic planning domains to satisfy: 
\textbullet~\emph{Existence of agent action}: $\forall s \in 2^\F.\exists a \in \alpha(s)$; \textbullet~\emph{Existence of environment reaction}: $\forall s \in 2^{\F}, a \in \alpha(s).\exists r \in \beta(s, a)$; \textbullet~\emph{Uniqueness of environment reaction}: $\forall s \in 2^{\F}, a \in \alpha(s).\delta(s, a, r_1) = \delta(s, a, r_2) \supset r_1 = r_2$.

These properties allow the defined nondeterministic planning domain to capture classical \FOND domains~\cite{CimattiRT98,GeBo2013} expressed in \PDDL~\cite{Haslum2019} by introducing reactions corresponding to \texttt{oneof} clauses of agent actions~\cite{DPZ23Ecai}.

A \emph{trace} of $\P$ is a finite or infinite sequence $\tau = s_0 (a_1, r_1, s_1) \cdots$, where $s_0$ is the initial state of $\P$, and $s_i$, $a_i$, and $r_i$, are the state, agent action, and environment reaction, respectively, at the $i$-th time step. 
A trace is \emph{legal} if for every $i > 0$: \myi $a_i \in \alpha(s_{i-1})$; \myii $r_i \in \beta(s_{i-1}, a_{i})$; and \myiii $s_{i} = \delta(s_{i-1}, a_{i}, r_{i})$. We denote by $\H_\P$ the set of legal traces of $\P$.
Given a trace $\tau = s_0 (a_1, r_1, s_1) \cdots (a_n, r_n, s_n)$, we denote $\tau$ projected on the states by $\tau|_{2^\F} = s_0 \cdots s_n$, and $\tau$ projected on agent actions and environment reactions by $\tau|_{\act \times \react} = (a_1, r_1) \cdots (a_n, r_n)$.




An \emph{agent strategy} is a \emph{partial} function $\sigma: (2^\F)^+ \rightarrow \act$ that maps sequences of states of the domain to agent actions.  We write $\sigma(\tau) = \bot$ to denote that $\sigma$ is undefined in $\tau$, which we think of as the strategy terminating its execution.
%
%
Note that such agent strategies definitions are more in line with existing works on planning and reasoning about actions. However, in synthesis agent strategies are typically defined as $\sigma': \react^* \rightarrow \act$ that map sequences of environment reactions to agent actions. However, for the planning domains we consider, we can equivalently define agent strategies as $\sigma:(2^\F)^+ \rightarrow \act$ or $\sigma':\react^* \rightarrow \act$. The equivalence follows by noting that: \myi every strategy $\sigma: (2^\F)^+ \rightarrow \act$ directly corresponds to a strategy $\sigma' : \react^* \rightarrow \act$ by the requirement of uniqueness of environment reaction; \myii every strategy $\sigma' : \react^* \rightarrow \act$ directly corresponds to a strategy $\sigma : (2^\F)^+ \rightarrow \act$ since the transition function $\delta$ is deterministic. 
An agent strategy $\sigma$ is \emph{legal} if, for every legal trace $\tau = s_0 (a_1, r_1, s_1) \cdots (a_n, r_n, s_n)$, if $\sigma(\tau|_{2^{\F}})$ is defined, then $\sigma(\tau|_{2^{\F}}) \in \alpha(s_n)$, so that a legal agent strategy always satisfies action preconditions. 

An \emph{environment strategy} is a \emph{total} function $\gamma: (\act)^+ \rightarrow \react$ mapping sequences of agent actions to environment reactions. An environment strategy $\gamma$ is \emph{legal} if, for every legal trace $\tau = s_0 (a_1, r_1, s_1) \cdots (a_n, r_n, s_n)$ and agent action $a_{n+1} \in \alpha(s_n)$, we have that $\gamma(a_1 \cdots a_{n+1}) \in \beta(s_n, a_{n+1})$, so that a legal environment strategy always satisfies reaction preconditions. In the rest of the paper, we consider only legal agent strategies and legal environment strategies. 

An agent strategy $\sigma$ and an environment strategy $\gamma$ induce a unique legal trace on $\P$ that is \emph{consistent} with both, defined as $\Play(\sigma, \gamma) = s_0 (a_1, r_1, s_1) \cdots$, where: \myi $s_0$ is the initial state; 
\myii for every $i > 0$, $a_i = \sigma(s_0 \cdots s_{i-1})$, $r_i = \gamma(a_1 \cdots a_{i})$, and $s_i = \delta(s_{i-1}, a_i, r_i)$; and \myiii if $\Play(\sigma, \gamma)$ is finite, say $\Play(\sigma, \gamma) = s_0(a_1, r_1, s_1) \cdots (a_n, r_n, s_n)$, then $\sigma(s_0 \cdots s_n) = \bot$. Note that $\Play(\sigma, \gamma)$ always exists by the properties of Existence of agent action and Existence of environment reaction. 

\paragraph{Winning and Cooperative Strategies.} Given a domain $\P$, an objective $\varphi$ as an \LTLf formula over the fluents in $\P$. A \emph{finite} legal trace $\tau$ satisfies $\varphi$ in $\P$, written $\tau \pmodels \varphi$, if $\tau|_{2^{\F}} \models \varphi$. We denote by $\L_{\P}(\varphi)$ the set of legal traces of $\P$ satisfying $\varphi$. An agent strategy $\sigma$ is \emph{cooperative} for $\varphi$ in $\P$ if $\Play(\sigma, \gamma)$ is finite and $\Play(\sigma, \gamma) \pmodels \varphi$ for some environment strategy $\gamma$. Furthermore, $\sigma$ is \emph{winning} for (or \emph{enforces}) $\varphi$ in $\P$ if $\Play(\sigma, \gamma)$ is finite and $\Play(\sigma, \gamma) \pmodels \varphi$ for every environment strategy $\gamma$.

\paragraph{\LTLftitle Best-Effort Synthesis.}~\label{sec:best-effort-synthesis} 
\LTLf best-effort synthesis in nondeterministic domains computes a strategy that enables the agent to do its best to satisfy an \LTLf objective in a nondeterministic domain~\cite{ADR21,DPZ23Ecai}.  

The notion of best-effort strategy bases on the game-theoretic relation of \emph{dominance}. An agent strategy $\sigma_1$ dominates an agent strategy $\sigma_2$ (for $\varphi$ in $\P$), written $\sigma_1 \geq_{\varphi|\P} \sigma_2$ if, for every environment strategy $\gamma$, $\Play(\sigma_2,\gamma)$ is finite and \(\Play(\sigma_2, \gamma) \models_{\P} \varphi\) implies $\Play(\sigma_1,\gamma)$ is finite and \(\mathtt{Play}(\sigma_1, \gamma) \models_{\P} \varphi\). Furthermore,  \(\sigma_{1}\) \textit{strictly dominates} \(\sigma_{2}\), written \(\sigma_1 >_{\varphi|\P} \sigma_2\), if \(\sigma_1 \geq_{\varphi|\P} \sigma_2\) and \(\sigma_2 \not \geq_{\varphi|\P} \sigma_1\).


Intuitively, \(\sigma_1 >_{\varphi|\P} \sigma_2\) means that \(\sigma_1\) does at least as well as \(\sigma_2\) against every environment strategy 
and strictly better against at least one such strategy. An agent using $\sigma_2$ is not doing its ``best'' to satisfy the goal. If the agent used $\sigma_1$ instead, it could achieve the goal against a strictly larger set of environment strategies. As a result, a best-effort strategy $\sigma$ is one that is not strictly dominated by any other strategy, i.e., no other strategy $\sigma'$ exists such that $\sigma' >_{\varphi|\P} \sigma$.


Best-effort strategies also admit an alternative characterization that describes their behavior when executed after a history, i.e., a finite legal trace. 
Given a domain $\P$, an agent strategy $\sigma$, and a history $h$, we denote by $\Gamma_{\P}(\sigma, h)$ the set of environment strategies $\gamma$ such that $h$ is a prefix of $\Play(\sigma, \gamma)$. Also, we denote by $\H_{\P}(\sigma)$ the set of histories $h$ such that $\Gamma_{\P}(\sigma, h)$ is non-empty, i.e., the set of histories that are consistent with $\sigma$ and some environment strategy $\gamma$.
Given an agent goal $\varphi$, we define: 
\begin{compactitem}
    \item 
    $\val{\varphi|\P}(\sigma, h) = \win$ ($\sigma$ is winning for $\varphi$ from $h$) 
    if $\Play(\sigma, \gamma)$ is finite and $\Play(\sigma, \gamma) \models_{\P} \varphi$ for every $\gamma \in \Gamma_{\P}(\sigma, h)$;
    \item 
    $\val{\varphi|\P}(\sigma, h) = \pend$  ($\sigma$ is pending for $\varphi$ from $h$)
    if, $\Play(\sigma, \gamma)$ is finite and $\Play(\sigma, \gamma) \models_{\P} \varphi$ for some $\gamma \in \Gamma_{\P}(\sigma, h)$, but not all;
    \item 
    $\val{\varphi|\P}(\sigma, h) = \lose$ ($\sigma$ is losing for $\varphi$ from $h$), otherwise. 
\end{compactitem}


The value of $h$, written $\val{\varphi|\P}(h)$\footnote{We define $\val{\varphi}(h)$ only if there exists $\sigma$ s.t. $h \in \H(\sigma)$.}, is the maximum of $\val{\varphi}(\sigma, h)$ for every agent strategy $\sigma$ such that $h \in \H_{\P}(\sigma)$.
The history-based characterization of best-effort strategies is as follows: an agent strategy $\sigma$ is best-effort for $\varphi$ in $\P$ iff $\val{\varphi|\P}(\sigma, h) = \val{\varphi|\P}(h)$ for every $h \in \H_{\P}(\sigma)$.

By the history-based characterization, 
a best-effort strategy $\sigma$ behaves as follows. Starting from every history $h \in \H_{\P}(\sigma)$: \myi  
if $\varphi$ is enforceable regardless of the nondeterminism in the domain, $\sigma$ enforces $\varphi$; else, \myii if $\varphi$ is satisfiable only depending on how the nondeterminism in the domain unfolds, $\sigma$  satisfies $\varphi$ if the environment cooperates; else, \myiii if $\varphi$ is unsatisfiable, $\sigma$ prescribes some action which does not violate action preconditions. 
Note that a best-effort strategy adapts so that if a goal becomes enforceable due to environment cooperation, then the strategy will enforce it.    

 By the definition of best-effort strategy, it follows that, if a winning strategy exists, the best-effort strategies are exactly the winning strategies. Furthermore, best-effort strategies have the notable property that they always exist. 
 \LTLf best-effort synthesis is \twoexptime-complete in the size of the goal $\varphi$ and \exptime-complete in the size of the domain $\P$~\cite{DPZ23}. 




\section{Multi-Tier Goals\label{sec:multi-tier-goals}}


In this paper, we address synthesis with multi-tier goals consisting of a multi-tier hierarchy of increasingly challenging \LTLf objectives:
the initial objective is the simplest, with each subsequent objective adding more obligations, making the final objective the most challenging.
Formally, we define a multi-tier goal specified in \LTLf as follows. 

\begin{definition}[Multi-Tier Goal]\label{def:multi-tier-goals}
    Given a nondeterministic planning domain $\P$ with fluents $\F$, a multi-tier \LTLf goal is a sequence of \LTLf objectives $\Phi = \langle \varphi_1, \cdots, \varphi_n \rangle$ over $\F$ such that $\L_{\P}(\varphi_1) \supseteq \cdots \supseteq \L_{\P}(\varphi_n)$. 
\end{definition}

Multi-tier \LTLf goals can capture \emph{ordinal temporal preferences}~\cite{son2006planning}, which enable users to specify when a strategy is better than another. 
Given a multi-tier goal \mbox{$\Phi = \langle \varphi_1, \cdots, \varphi_n \rangle$}, we can view $\varphi_1$ as the ``hard'' objective that the agent should enforce, and the various $\varphi_i$~(where $i > 1$) as ``soft'' objectives that refine $\varphi_1$, which the agent should satisfy if possible~\cite{keyder2009soft}. As a result, the strategies that satisfy $\varphi_i$ are preferred to those that only satisfy $\varphi_j$ for $j < i$. 

It follows by Definition~\ref{def:multi-tier-goals} that if an agent strategy enforces~(resp. cooperates for) $\varphi_i$, then it also enforces (resp. cooperates for) every $\varphi_j$ such that $j < i$. 


One important aspect of synthesis in nondeterministic domains is that, at every time step, an objective $\varphi$ can be: \begin{compactitem}
    \item \emph{Enforceable}, i.e., there exists an agent strategy that satisfies $\varphi$ 
    regardless of adversarial environment responses;
    \item \emph{Pending}, i.e., there exists an agent strategy that satisfies $\varphi$ 
    should the environment cooperatively respond;
    \item \emph{Unsatisfiable}, i.e., no agent strategy exists that satisfies $\varphi$
    regardless of environment responses.
\end{compactitem}

It is worth noting that an objective that is currently pending might become enforceable during execution should the environment cooperate (or unsatisfiable should the environment be adversarial). 
Hence in order to fully exploit the nondeterminism in a nondeterministic domain considering a multi-tier goal, the agent should employ an \emph{adaptive strategy}, i.e., a strategy with the following properties:
\begin{compactitem}
    \item \emph{Property 1.} The strategy enforces the satisfaction of all objectives that are currently enforceable;
    \item \emph{Property 2.} The strategy exploits possible cooperation from the environment to satisfy as many as possible of the currently pending objectives;
    \item \emph{Property 3}. Dynamically, at each time step, if the environment cooperates and a pending objective becomes enforceable, the strategy will enforce its satisfaction. 
\end{compactitem}


Synthesizing a strategy that achieves Properties 1, 2, and 3, requires more sophisticated techniques than those developed for PBP in deterministic domains~\cite{son2006planning,bai-bac-mci-aij09,keyder2009soft}, where the environment response is fixed and objectives are only either enforceable or unsatisfiable.
Moreover handling ordinal temporal objectives in \FOND requires significant advancements compared to PBP in \FOND planning addressed in~\cite{shaparau2006contingent} for static preferences (which can only express additional ``soft'' reachability objectives). In particular, the machinery developed for best-effort strategies for singular objectives \cite{ADR21} plays a prominent role in synthesizing adaptive strategies for multi-tiers goals.

\section{\LTLf Adaptive Synthesis for \\ Multi-Tier Goals~\label{sec:problem}}

We now formalize the notion of adaptive strategies for multi-tier goals that comply with \mbox{Properties 1, 2, and 3} discussed in the previous section.


We define adaptive strategies for a multi-tier goal by leveraging the notion of best-effort strategy for a single objective. 
Specifically, we leverage the history-based characterization of best-effort strategies to satisfy the three properties above when considering multi-tier goals. 


Note that an adaptive strategy requires identifying all enforceable objectives at every point in time~(cf. Property 1 and Property 3). Therefore, we define the \emph{maximally winning objective} as the highest tier that the agent can enforce, regardless of an adversarial environment in a nondeterministic planning domain, for any given history.

\begin{definition}[Maximally Winning Objective]\label{def:def:maximally-winning-goals}
Let $\P$ be a nondeterministic planning domain, $\Phi = \langle \varphi_1, \cdots, \varphi_n \rangle$ a multi-tier goal, and $h \in \H_\P$ a history. 
We say that $\varphi_\ell$ is the maximally winning objective 
wrt $h$ if $\val{\varphi_\ell|\P}(h) = \win$ and $\val{\varphi_{j}|\P}(h) \neq \win$ for every $j > \ell$.
\end{definition}

One can define the \emph{maximally pending objective} analogously, i.e., the highest tier that the agent can satisfy should the environment cooperate.

An adaptive strategy must always leverage possible environment cooperation to fulfill all objectives that are currently not enforceable but satisfiable, while continuously enforcing those objectives that are enforceable. To this end, we define the \emph{maximally winning-pending objective}, referring to the highest tier objective the agent can satisfy with environment cooperation, while simultaneously enforcing those currently enforceable~(cf. Property 1 and Property 2).

\begin{definition}[Maximally Winning-Pending Objective]\label{def:maximally-winning-pending-goals}
    Let $\P$ be a nondeterministic planning domain, $\Phi = \langle \varphi_1, \cdots, \varphi_n \rangle$ a multi-tier goal, $h \in \H_\P$ a history, and $\varphi_\ell$ the maximally winning objective of $\Phi$ 
    wrt $h$.
    We say that $\varphi_{k}$ ($k > \ell$) is the maximally winning-pending objective 
    wrt $h$ if \myi there exists a strategy $\sigma$ consistent with $h$ that is winning for $\varphi_\ell$ such that $\val{\varphi_{k}|\P}(\sigma, h) = \pend$, and \myii it does not exist a strategy $\eta$ consistent with $h$ and winning for $\varphi_\ell$ such that $\val{\varphi_j|\P}(\eta, h) = \pend$ for some $j > k$.
\end{definition}


We now define our notion of adaptive strategy for multi-tier goals. Specifically:
\begin{compactitem}
    \item To achieve Property 1, the adaptive strategy will always enforce the maximally winning objective, if one exists, regardless of the adversarial environment, thus enforcing every objective that is currently enforceable; 
    \item To achieve Property 2, the adaptive strategy will always satisfy the maximally winning-pending objective or maximally pending objective~(if the maximally winning objective does not exist), if one exists, should the environment cooperate, thus satisfying every pending objective;
    \item To achieve Property 3, Properties 1 and 2 should hold for every history consistent with the adaptive strategy. 
\end{compactitem}

Formally, adaptive strategies are defined as follows:

\begin{definition}[Adaptive Strategy for Multi-Tier Goals]~\label{def:best-effort-plan}
    Let $\P$ be a nondeterministic planning domain and $\Phi = \langle \varphi_1, \cdots, \varphi_n \rangle$ a multi-tier goal. An agent strategy $\sigma$ is an adaptive strategy for $\Phi$ in $\P$ if, for every $h \in \H_{\P}(\sigma)$, one of the following holds: 
    \begin{enumerate}[noitemsep,topsep=0pt,parsep=0pt,partopsep=0pt]
        \item 
        Suppose both the maximally winning objective $\varphi_{\ell}$ and the maximally winning-pending objective $\varphi_k$
        exist, then: \myi $\sigma$ is winning for $\varphi_\ell$ in $\P$ from $h$, and \myii there exists an agent strategy $\sigma'$ such that $\val{\varphi_k|\P}(\sigma', h) = \pend$ and $\sigma(h) = \sigma'(h)$ 
        \item 
        Suppose the maximally winning objective  $\varphi_{\ell}$ 
        exists, yet no maximally winning-pending 
        exists, then $\sigma$ is winning for $\varphi_{\ell}$ in $\P$ from $h$;
        \item 
        Suppose neither the maximally winning objective nor the maximally winning-pending objective exists, yet there exists a maximally pending objective $\varphi_{p}$, then there exists an agent strategy $\sigma'$ such that $val_{\varphi_p|\P}(\sigma', h) = \pend$ and $\sigma(h) =  \sigma'(h)$. 
    \end{enumerate}
\end{definition}

In this paper, we study synthesis of adaptive strategies for \LTLf multi-tier goals in nondeterministic planning domains. 

\begin{definition}[\LTLf Adaptive Synthesis for Multi-Tier Goals]~\label{def:ltlf-best-effort-synth}
    Given a nondeterministic planning domain $\P$ and an \LTLf multi-tier goal $\Phi$, adaptive synthesis for multi-tier goals is to compute an adaptive strategy for $\Phi$ in $\P$. 
\end{definition}

The major contribution of this paper is a game-theoretic synthesis technique that computes adaptive strategies for \LTLf multi-tier goals, which we will show in Section~\ref{sec:technique}. The correctness of the synthesis technique also shows that \LTLf adaptive strategies for multi-tier goals always exist~(analogously to \LTLf best-effort strategies for single objectives). 

\begin{theorem}~\label{thm:existence}
    An adaptive strategy always exists for an \LTLf multi-tier goal $\Phi$ in a planning domain $\P$.
\end{theorem}

\section{Illustration Example.\label{sec:example}}
We present an example drawn from robot navigation 
to illustrate the notions presented in previous sections and show how multi-tier goals can capture temporal preferences.   

\begin{figure}[t!]
    \centering
    \includegraphics[width=0.705\linewidth]{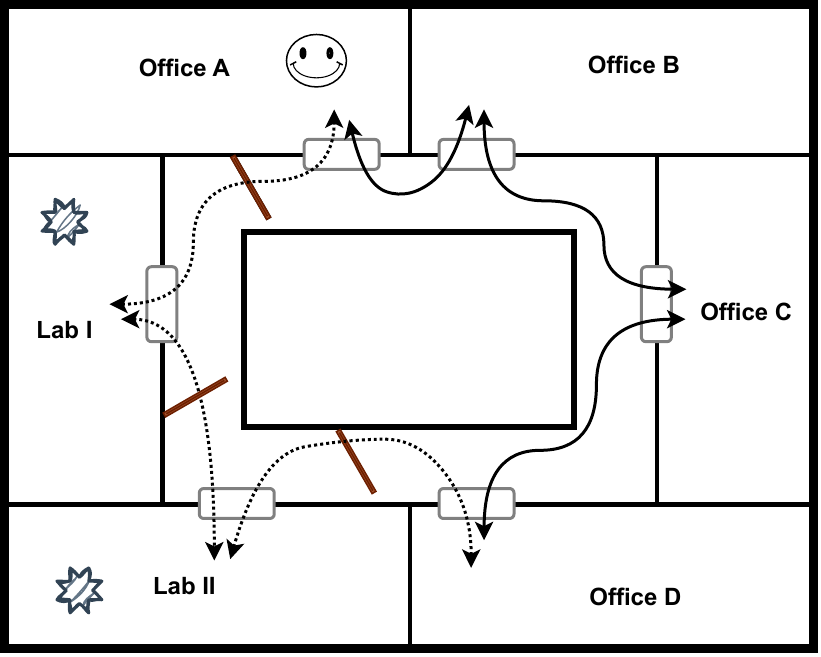}
    \caption{Robot working in an office building.}
    \label{fig:example}
\end{figure}%


Consider a cleaning robot operating in a circular building, as shown in Figure~\ref{fig:example}. The robot can freely access Offices A, B, C, and D. However, entry to Labs I and II requires passing through secure gates in the hallway, shown in brown in Figure~\ref{fig:example}. These gates restrict access when hazardous materials are present, managed by the building manager. Solid arrows indicate areas with unrestricted access, while dashed arrows represent limited-access zones. Assume the robot starts in Office A, with the secure gates initially open. Consider the multi-tier goal $\Phi = \langle \varphi_1, \varphi_2, \varphi_3 \rangle$, where:
\begin{compactitem}
    \item $\varphi_1 = \Diamond (\textit{Office\_D\_clean})$;
    \item $\varphi_2 = \Diamond(\textit{Office\_D\_clean}) \land \Diamond(\textit{Lab\_II\_clean})$; 
    \item $\varphi_3 = \Diamond(\textit{Lab\_II\_clean} \land \Next(\Diamond(\textit{Office\_D\_clean})))$; 
\end{compactitem}

The multi-tier goal $\Phi$ specifies the robot should clean Office D ($\varphi_1$), but cleaning Lab II also is preferable ($\varphi_2$), and cleaning Lab II before Office D ($\varphi_3$) is even more preferable. 

Initially, $\varphi_1$ is enforceable, while $\varphi_2$ and $\varphi_3$ are pending. Thus, $\varphi_1$ is the maximally winning objective at this step. A winning strategy for $\varphi_1$ is to clean Office D without selecting a path that traverses the secure gates.
This is because if the robot moves through the secure gates, the building manager might close the gates and lock the robot in one of the Labs, thus preventing $\varphi_1$ from being satisfied. 

With $\varphi_1$ as the maximally winning objective, we have that $\varphi_2$ is the maximally winning-pending objective, as any winning strategy for $\varphi_1$ avoids clean Lab II before Office D such that not satisfying $\varphi_3$. A strategy $\sigma$ that navigates through Offices A, B, C, and D, cleans Office D, and proceeds to cleans Lab II if the secure gate is open, is winning for $\varphi_1$ and cooperative for $\varphi_2$. 
Once the robot finishes cleaning Office D, $\varphi_2$ becomes enforceable if the building manager left secure gates open, in which case $\varphi_2$ becomes the maximally winning objective and $\sigma$ enforces $\varphi_2$ as well. 
Thus $\sigma$ is a strategy that satisfies Definition~\ref{def:best-effort-plan} and is adaptive for $\Phi$.

This example also shows that a best-effort strategy for any single objective in $\Phi$ is not guaranteed to be an adaptive strategy for the multi-tier goal $\Phi$.
A best-effort (winning) strategy for $\varphi_1$ can navigate to Offices A, B, C, and D, then clean Office D, and stay there, so that $\varphi_2$ would never be satisfied. Instead, a best-effort strategy for $\varphi_2$ can navigate to and clean Lab II and thereafter navigate to and clean Office D. This might get the robot stuck between secure gates so that not satisfying $\varphi_1$~(analogously for $\varphi_3$).

\section{Building Blocks~\label{sec:games}}

We develop a game-theoretic technique to address adaptive synthesis for multi-tier goals, involving two-player games over deterministic finite automata, reviewed briefly below.   


A \emph{deterministic transition system} is a tuple $\T = (\Sigma, Q, \iota, \varrho)$, where: $\Sigma$ is a finite input alphabet; $Q$ is a finite set of states; $\iota \in Q$ is the initial state; and $\varrho: Q \times \Sigma \rightarrow Q$ is the transition function. The \emph{size} of $\T$ is $|Q|$.
Given a finite trace $\pi = \pi_0 \cdots \pi_n$ over $\Sigma$, we extend $\varrho$ to a function $\varrho: Q \times \Sigma^* \rightarrow Q$ as follows:
$\varrho(q, \lambda) = q$ 
and, if $q_n = \varrho(q, \pi_0 \cdots \pi_{n-1})$, then $\varrho(q, \pi_0 \cdots \pi_n) = \varrho(q_n, \pi_n)$.

\begin{definition}~\label{def:synch-product}
    The product of two transition systems $\T_i = (\Sigma, Q_i, \iota_i, \varrho_i)$ (for $i = 1, 2$) is the transition system $\textsc{Product}(\T_1, \T_2) =$ $ (\Sigma, Q_1 \times Q_2, (\iota_1, \iota_2), \varrho)$, where $\varrho((q_1, q_2), a, r) = (\varrho_1(q_1, a, r), $ $\varrho_2(q_2, a, r))$.
\end{definition}


Given a transition system $\T = (\Sigma, Q, \iota, \varrho)$ and a set of states $V \subseteq Q$, the complement of $V$ wrt $Q$ is $\overline{V} = Q \setminus V$.

A \emph{deterministic finite automaton} (\DFA) is a tuple $\A = (\T, R)$, where $\T = (\Sigma, Q, \iota, \varrho)$ is a transition system, and $R \subseteq Q$ is a set of \emph{final states}.  
A word $\trace \in \Sigma^*$ is \emph{accepted} by $\A$ if $\varrho(\iota, \trace) \in R$. The language recognized by $\A$, written $\L(\A)$, is the set of words that the automaton accepts. 

\begin{theorem}\cite{DegVa13}\label{thm:ltlf-to-nfa-dfa}
    Given an \LTLf formula $\varphi$ we can build a \DFA, denoted $\textsc{ToDFA}(\varphi)$, with size at most double-exponential in $|\varphi|$ and whose language is the set of finite traces that satisfy $\varphi$.
\end{theorem}

A \DFA game is a \DFA with alphabet $\act \times \react$, where $\act$ and $\react$ are two disjoint sets under control of agent and environment, respectively. Formally, a \DFA game is a pair $\G = (\T, R)$, where: $\T = (\act \times \react, Q, \iota, \varrho)$ is a transition system
and $R \subseteq Q$ is a set of final states. 

A game strategy is a partial function $\kappa: Q \rightarrow \act$ that maps states of the game to agent actions. Given a game strategy $\kappa$, a sequence of environment reactions $\vec{r} = r_0 r_1 \ldots \in \react^\omega$, and a \DFA game $\G$, we denote by $\run(\kappa, \vec{r}, \G)$, the \emph{run} $q_0 q_1 \cdots$ of states of $\G$ \emph{induced} by~(or \emph{consistent with}) $\kappa$ and $\vec{r}$ as follows: \myi $q_0$ is the initial state of $\G$; \myii for every $i \geq 0$, we have $q_{i+1} = \varrho(q_{i}, a_{i}, r_{i})$, where $a_i = \kappa(q_i)$; and \myiii if $\run(\kappa, \vec{r}, \G) = q_0 \cdots q_n$ is finite, then $\kappa(q_n) = \bot$.
A game strategy is winning (resp. cooperative) in $\G$ if $\rho = \run(\kappa, \vec{r}, \G)$ is finite and $\last(\rho) \in R$ for every $\vec{r} \in \react^\omega$ (resp. for some $\vec{r} \in \react^\omega$). 
A state $q \in Q$ is a \emph{winning} (resp. \emph{cooperative}) \emph{state} if the agent has a winning (resp. cooperative) game strategy in the game $\G' = (\T', R)$, where $\T' = (\act\times\react, Q, q, \varrho)$, i.e., the same game as $\G$, but with initial state $q$. The \emph{winning} (resp. \emph{cooperative}) \emph{region} $\w$ (resp. $\c$) is the set of winning (resp. cooperative) states. A game strategy that is winning from every state in the winning (resp. cooperative) region is called \emph{uniform winning} (resp. \emph{uniform cooperative}). We observe that while a game strategy is not formally an agent strategy (i.e., a function from environment reactions to agent actions), it \emph{induces} one as follows.

\begin{definition}~\label{dfn:induced-strategy}
    Given a transition system $\T = (\act \times \react, Q, \iota, \delta)$, a game strategy $\kappa: Q \rightarrow \act$ induces an agent strategy $\sigma':\react^* \rightarrow \act$ as follows: for every $h \in (\act \times \react)^*$, $\sigma'(h|_{\react}) = \kappa(\delta(\iota,h))$. The pair $(\T, \kappa)$ denotes a \emph{transducer}, i.e., a transition system with output function $\kappa$.
    \end{definition}
\noindent (We also recall: in our setting every strategy $\sigma': \react^* \rightarrow \act$ is equivalent to a strategy $\sigma: (2^{\F})^+\rightarrow\act$.)

\emph{Solving} a \DFA game $\G$ aims at computing the winning (resp. cooperative) region and a uniform winning (resp. cooperative) strategy, written $(\w, \kappa) = \textsc{SolveAdv}(\G)$ (resp. $(\c, \nu) = \textsc{SolveCoop}(\G)$). \DFA games can be solved in linear time in their size by a fixpoint computation over the state space of the game~\cite{AG11}. For states 
not in $\w$ (resp. $\c$), we assume that $\kappa$ (resp. $\nu$) is undefined.

Given two \DFAs $\A_i = (\T_i, R_i)$ for $(i = 1, 2)$ and the product $\T$ of their transition systems, we can \emph{lift} the final states $R_i$ to the product $\T$ as follows.

\begin{definition}
    For \DFAs $\A_i = (\T_i, R_i)$ and $\T = \textsc{Product}(\T_1, \T_2)$, the lifting of $R_i$ to $\T$ is the set of states $\textsc{Lift}(\T, R_i) = \{(q_1, q_2) \in Q_1 \times Q_2 \tst q_i \in R_i\}$.
\end{definition}

We now show how to transform planning domains into \DFAs by introducing two error states $\agerr$ and $\enverr$, denoting that agent and environment violate their respective preconditions, as in~\cite{DPZ23}. 


\begin{definition}~\label{def:domain-2-dts}
    Given a nondeterministic planning domain $\P = (2^\F, s_0, \act, \react, \alpha, \beta, \delta)$, we define $(\P_+, \{\agerr\}, \{\enverr\}) = \textsc{ToDFA}(\P)$\footnote{For simplicity, we extend the notation \textsc{ToDFA} to return two DFAs $(\P_+, \{\agerr\}, \{\enverr\})$ and $(\P_+, \{\agerr\}, \{\enverr\})$, which share the same transition system but differ in their final states.}, 
    where $\P_+ = (\act \times \react, 2^{\F} \cup \{\agerr, \enverr\}, s_0, \delta')$ is a transition system with transition function $\delta'$ such that $\delta'(s, a, r) = \delta(s, a, r)$ if $a \in \alpha(s)$ and $r \in \beta(s, a)$; or $\delta'(s, a, r) = s^{ag}_{err}$ if $a \not \in \alpha(s)$; or $\delta'(s, a, r) = s^{env}_{err}$ if $a \in \alpha(s)$ and $r \not \in \beta(s, a)$.  
\end{definition}

Our synthesis technique utilizes the synthesis approach developed for \LTLf synthesis in nondeterministic domains~\cite{DPZ23Ecai}. This approach involves constructing the product of the transition systems of the domain, defined over $\act \times \react$ and the \DFA of the \LTLf objective, defined over $2^{\F}$.  


\begin{definition}\label{def:product-2}
    Let $\P_+ = (\act \times \react, 2^{\F} \cup \{\agerr, \enverr\}, s_0, \delta')$ be the transition system of a domain $\P$ and $\T_\varphi = (2^\F, Q, \iota, \varrho)$ the transition system of the \DFA of an \LTLf objective. The product of $\P_+$ and $\T_{\varphi}$ is $\textsc{Product}(\P_+, \T_{\varphi}) = (\act \times \react, (2^{\F} \cup \{\agerr, \enverr\}) \times Q, (s_0, \varrho(\iota, s_0)), \partial)$, where: \\
        $\partial((s, q), a, r) =
      \begin{cases}
    (s', \varrho(q, s')) &\text{if } s' \not \in \{s^{ag}_{err}, s^{env}_{err}\} \\
    (s^{ag}_{err}, q) &\text{if }  s' = s^{ag}_{err}\\
    (s^{env}_{err}, q) &\text{if } s' = s^{env}_{err}
    \end{cases}
    $
     
    \noindent with $s' = \delta'(s, a, r)$.

\end{definition}

Intuitively, $\T$ is a transition system that simultaneously retains the progression in $\P$ and $\T_\varphi$. 
\section{Synthesis Technique~\label{sec:technique}}
In this section, we present a game-theoretic technique to address adaptive synthesis for \LTLf multi-tier goals in nondeterministic planning domains. This technique relies on solving and combining on-the-fly the solutions of several \DFA games, constructed from both the planning domain and the objectives within the multi-tier goal. 

We first review the key steps for best-effort synthesis of a single \LTLf objective~\cite{DPZ23}, as outlined in Algorithm~\ref{alg:single-goal-be}.  
That solves two distinct games over the same transition system $\T$ (with different final states) constructed from the domain and the \LTLf objective. 
Solving these games returns the following: the winning region $\w$ with a uniform winning game strategy $\kappa$, and the cooperative region $\c$ with a uniform cooperative game strategy $\nu$. 
We have that $\w$, $\kappa$, $\c$, and $\nu$ satisfy the following: 
paths ending in $\w$ correspond
to histories with value $\win$, as witnessed by $\kappa$; 
paths ending in $\c\setminus\w$ correspond to histories with value $\pend$, as witnessed by $\nu$; the remaining paths correspond to histories with value $\lose$. 
This property will later serve to detect maximally winning and maximally pending objectives when synthesizing adaptive strategies.



\begin{algorithm}[t]
\caption{$\textsc{SynthDSSingleObj}(\P, \varphi)$}\label{alg:single-goal-be}
\begin{algorithmic}[1]
\REQUIRE Domain $\P$ and \LTLf goal $\varphi$
\ENSURE Trans. sys. $\T$, win. region $\w$, win. strategy $\kappa$, coop. region $\c$, and coop. strategy $\nu$ 
\STATE ($\P_+, \{\agerr\}, \{\enverr\}) = \textsc{ToDFA}(\P)$
\STATE $(\T_{\varphi}, R_{\varphi}) = \textsc{ToDFA}(\varphi)$
\STATE $\T = \textsc{Product}(\P_+, \T_{\varphi})$ \\
\STATE $\bigagerr = \textsc{Lift}(\T, \{\agerr\})$ 
\STATE $\bigenverr = \textsc{Lift}(\T, \{\enverr\})$
\STATE $R'_{\varphi} = \textsc{Lift}(\T, R_{\varphi})$
\STATE $Adv = \overline{\bigagerr} \cap (\bigenverr \cup R'_{\varphi})$
\STATE $Coop = \overline{\bigagerr} \cap \overline{\bigenverr} \cap R'_{\varphi}$
\STATE $(\w, \kappa) = \textsc{SolveAdv}(\T, Adv)$
\STATE $(\c, \nu) = \textsc{SolveCoop}(\T, Coop)$
\STATE \textbf{Return} $(\T, \w, \c, \kappa, \nu)$ 
 \end{algorithmic} 
\end{algorithm}


\begin{algorithm}[t]
    \caption{$\textsc{SynthDSWinPend}(\P, \varphi_1, \varphi_2)$}\label{alg:strong-weak-plan}
    \begin{algorithmic}[1]
        \REQUIRE Domain $\P$; \LTLf objs $\varphi_1, \varphi_2$ st $\L_{\P}(\varphi_1) \supseteq \L_{\P}(\varphi_2)$%
        \ENSURE Win-coop region $\wc$; agent strategy $\sigma$
        \STATE ($\P_+, \agerr, \enverr) = \textsc{ToDFA}(\P)$
        \STATE \textbf{For} $i = 1, 2$: \begin{compactenum}
            \item[2.1:] $(\T_{\varphi_i}, R_{\varphi_i}) = \textsc{ToDFA}(\varphi)$
            \item[2.2:] $\T_i = \textsc{Product}(\P_+, \T_{\varphi_i})$ \\
            /* $Q'_i$ is the state space of $\T_i$ */
            \item[2.3:] $\bigagerr_i = \textsc{Lift}(\T_i, \{\agerr\})$ 
            \item[2.4:] $\bigenverr_i = \textsc{Lift}(\T_i, \{\enverr\})$
            \item [2.5:] $R'_{\varphi} = \textsc{Lift}(\T, R_{\varphi})$
            \item[2.6:] $Adv_i = \overline{\bigagerr_i} \cap (\bigenverr_i \cup R'_{\varphi_i})$
            \item[2.7:] $Coop_i = \overline{\bigagerr_i} \cap \overline{\bigenverr_i} \cap R'_{\varphi_i}$
        \end{compactenum}
        \STATE $(\w_1, \kappa_1) = \textsc{SolveAdv}(\T_1, Adv_1)$
        \STATE $(\c_2, \text{-}) = \textsc{SolveCoop}(\T_2, Coop_2)$
        \STATE $\T = \textsc{Product}(\T_1, \T_2)$ \\
        ~~/* $Q' = Q'_1 \times Q'_2$ is the state space of $\T$ */ \\
        ~~/* $\partial$ is the transition function of $\T$ */
        \STATE Let $\wc_{-1} = \{\emptyset\}$ and $\wc_{0} = Adv_1 \times Coop_2$
        \STATE While $\wc_{i+1} \neq \wc_{i}$: \\
            ~~Let $\wc_{i+1} = \{q' \in Q' \ | \ \exists a.\exists r.\partial(q', a, r) \in \wc_{i} \land$ \\ $~~~~~~~~~~~~~~~~\forall r.\partial(q', a, r) \in \wc_{i} \cup (\w_1 \times \overline{\c_2})\}$ 
        \STATE Define strategy $\sigma$ on $\T$ as follows. For every $(q'_1, q'_2)$: \\ 
        $\sigma((q'_1, q'_2)) = $  $\begin{cases}
            a & \text{ if } (q'_1, q'_2) \in \wc_{i+1} \setminus \wc_{i} \\
            & \text{ and } \exists r.\partial((q'_1, q'_2), a, r) \in \wc_{i} \\
            \kappa_1(q'_1) & \text{ otherwise}
        \end{cases}$
        \STATE \textbf{Return} $(\wc, \sigma)$
    \end{algorithmic}
\end{algorithm}

A crucial step in adaptive synthesis for multi-tier goals is to determine, for every history, the current maximally winning-pending objective.
To this end, we provide in Algorithm~\ref{alg:strong-weak-plan} an auxiliary procedure that, for a pair of \LTLf objectives $\langle \varphi_1, \varphi_2 \rangle$ such that $\L_{\P}(\varphi_{1}) \supseteq \L_{\P}(\varphi_{2})$, computes a strategy $\sigma$ that satisfies the following: for every history $h$, 
%
if there exists a strategy to enforce $\varphi_1$ from $h$ and to satisfy $\varphi_2$ from $h$ with environment cooperation, i.e., currently, $\varphi_1$ is a winning objective and $\varphi_2$ is a winning-pending objective, then $\sigma$ should fulfill both $\varphi_1$ and $\varphi_2$ if the environment cooperates; if fulfilling both is not feasible, i.e., currently, $\varphi_2$ is not a winning-pending objective, then $\sigma$ enforces only $\varphi_1$. 
Intuitively, $\sigma$ prioritizes satisfying both $\varphi_1$ and $\varphi_2$ when possible, but defaults to enforcing $\varphi_1$ if necessary. 
%


Concretely, Algorithm~\ref{alg:strong-weak-plan} first computes the following: 
the winning region $\w_1$ and a winning strategy $\kappa_1$ for the game obtained from $\varphi_1$; and the cooperative region $\c_2$ for the game obtained from $\varphi_2$ (Lines~1-4).
Next, Algorithm~\ref{alg:strong-weak-plan} identifies game states, where paths ending at these states indicate that currently $\varphi_1$ is the maximally winning objective and $\varphi_2$ is the maximally winning-pending objective within the multi-tier goal $\langle \varphi_1, \varphi_2 \rangle$.
Such states are collected in $\wc$ through the fixpoint computation in Lines~5-7, in a game that aims to satisfy $\varphi_1$ adversarially, meanwhile satisfying $\varphi_2$ cooperatively.
A state $q'$ is added to $\wc_{i+1}$ when there exists an agent action $a$ such that: \myi some environment reaction moves the agent towards satisfying both $\varphi_1$ and $\varphi_2$, written $\exists r.\partial(q', a, r) \in \wc_{i}$; \emph{and} \myii every environment reaction either moves the agent towards satisfying both $\varphi_1$ and $\varphi_2$ or prevents $\varphi_2$ from being satisfied, but still ensures $\varphi_1$, written $\forall r.\partial(q', a, r) \in \wc_{i} \cup (\w_1 \times \overline{\c_2})$.
Paths ending in $\wc$ directly corresponds to histories that admit a strategy that simultaneously wins $\varphi_1$ and cooperates for $\varphi_2$.
Algorithm~\ref{alg:strong-weak-plan} constructs the output strategy $\sigma$ by combining $\kappa_1$ for enforcing $\varphi_1$ by default and the fixpoint computation information while collecting $\wc$~(Line~8). 
The construction is the following: for every state in $\wc$, 
for which there exists an action that definitely advances $\varphi_1$ and can advance $\varphi_2$ if the environment cooperates, then $\sigma$ follows this action~(first case, Line~8); otherwise $\sigma$ follows $\kappa_1$, advancing $\varphi_1$ only~(second case, Line~8).



Let $\val{\langle \varphi_1, \varphi_2 \rangle|\P}(h)$ denote $\langle \val{\varphi_1|\P}(h), \val{\varphi_2|\P}(h)\rangle$ (and similarly for $\val{\langle \varphi_1, \varphi_2 \rangle|\P}(\sigma, h)$). The following theorem shows the correctness of Algorithm~\ref{alg:strong-weak-plan}.

\begin{theorem}~\label{thm:win-pend-strategy}
    Let $\P$ be a domain, $\varphi_1$ and $\varphi_2$ two \LTLf objectives such that $\L_{\P}(\varphi_{1}) \supseteq \L_{\P}(\varphi_{2})$, and $\sigma$ the strategy returned by Algorithm~\ref{alg:strong-weak-plan}. For every $h \in \H_{\P}(\sigma)$, if $\val{\langle \varphi_1,\varphi_2 \rangle|\P}(h) = \langle \win, \pend \rangle$, either: \begin{enumerate}[noitemsep,topsep=0pt,parsep=0pt,partopsep=0pt]
        \item 
        $\val{\langle \varphi_1, \varphi_2 \rangle|\P}(\sigma, h) = \langle \win, \pend \rangle$; or
        \item 
        If no agent strategy $\eta$ exists s.t. $\val{\langle \varphi_1, \varphi_2 \rangle|\P}(\eta, h) = \langle \win, \pend \rangle$, then $\val{\varphi_1|\P}(\sigma, h) = \win$.
    \end{enumerate}
\end{theorem}

\begin{algorithm}[t]
\caption{$\textsc{SynthMultiTier}(\P, \Phi)$}\label{alg:super-mega-iper-ultra-multi-tier}
    \begin{algorithmic}[1]
        \REQUIRE Domain $\P$; Multi-tier goal $\Phi = \langle \varphi_1, \cdots, \varphi_n \rangle$.
        \ENSURE Agent strategy that is adaptive for $\Phi$ in $\P$
        \STATE \textbf{For} $i = 1, \cdots, n$: \begin{compactenum}
            \item[1.1:] \mbox{$(\T_i, \w_i, \c_i, \kappa_i, \nu_i) = \textsc{SynthDSSingleObj}(\P, \varphi_i)$;} \\
            \item[1.2:] \textbf{For} $j = i+1, \cdots, n$: \\
                $(\wc_{ij}, \omega_{ij}) = \textsc{SynthDSWinPend}(\P, \varphi_i, \varphi_j)$
        \end{compactenum}
        \STATE \textbf{Return} $\sigma: (2^\F)^+ \rightarrow \act$ defined as follows. \\
        Let $h$ be the input history: 
        \begin{compactenum}
            \item[$\bullet$] 
            \textbf{For} $i = 1, \cdots, n$: let $q'_i = \partial_i(\iota_i, h|_{\act\times\react})$, where $\iota_i$ and $\partial_i$ are the initial state and transition function of $\T_i$, respectively. 
            \item[$\bullet$] 
            $j = \textit{max}\{i \tst q'_i \in \w_i\}$ 
            \item[$\bullet$]
            $\ell = \textit{max}\{i \tst j > i \text{ and } (q'_i, q'_j) \in \wc_{ij}\}$
            \item[$\bullet$] 
            $m = \textit{max}\{i \tst q'_i \in \c_i\}$ 
        \end{compactenum}
        $\sigma(h) = \begin{cases}
            \omega_{j \ell}(q'_j, q'_\ell) &\text{ if } \ell > j > 0 \text{ exists} \\
             \kappa_{j}(q'_j) &\text{ else if } j > 0 \text{ exists}\\
             \nu_{m}(q'_m) &\text{ else if } m > 0 \text{ exists}\\
             \bot &\text{ otherwise}
        \end{cases}
        $
    \end{algorithmic}
\end{algorithm}

With Algorithms~\ref{alg:single-goal-be} and \ref{alg:strong-weak-plan} in place, we present our complete adaptive synthesis technique for \LTLf multi-tier goals, 
outlined in Algorithm~\ref{alg:super-mega-iper-ultra-multi-tier}.
Algorithm~\ref{alg:super-mega-iper-ultra-multi-tier} begins by computing all the winning and cooperative game strategies for every objective in the multi-tier goal $\Phi$ using Algorithm~\ref{alg:single-goal-be}. It then proceeds to compute strategies for winning-pending objectives with Algorithm~\ref{alg:strong-weak-plan}, covering all combinations of objective pairs ($\varphi_i, \varphi_j$), where $j > i$. 
The output adaptive strategy is generated in the form of an executable program that selects, at each time step, the next action on-the-fly from among the actions suggested by all the previously computed strategies.
The selection procedure involves checking the state corresponding to the input history $h$ to identify which objectives are currently winning, pending, and winning-pending.
Recall that paths ending in $\w_i$~(resp. $\c_i$) are with value $\win$~(resp. $\pend$). Hence we can determine the maximally winning objective $\varphi_j$, where $j = \textit{max}\{i \tst q'_i \in \w_i\}$. Subsequently, we can determine the maximally winning-pending objective $\varphi_\ell$, where $\ell = \textit{max}\{i \tst j > i \text{ and } (q'_i, q'_j) \in \wc_{ij}\}$. Analogously for the maximally pending objective  $\varphi_m$.
Based on this, the strategy selects an action following Properties 1,2\&3.
In details, at each time step, the strategy selects the output action as follows: \myi it selects the action returned by the strategy $\omega_{j\ell}$ that enforces the maximally winning objective $\varphi_j$ and cooperates for the maximally winning-pending objective $\varphi_\ell$, if any; else \myii it selects the action returned by the strategy $\kappa_j$ that enforces the maximally winning objective, if any; else \myiii it selects the action returned by the strategy $\nu_m$ that cooperates for the maximally pending objective $\varphi_m$, if any; else \myiv is undefined.  

\begin{figure}
    \centering
    \includegraphics[width=0.98\linewidth]{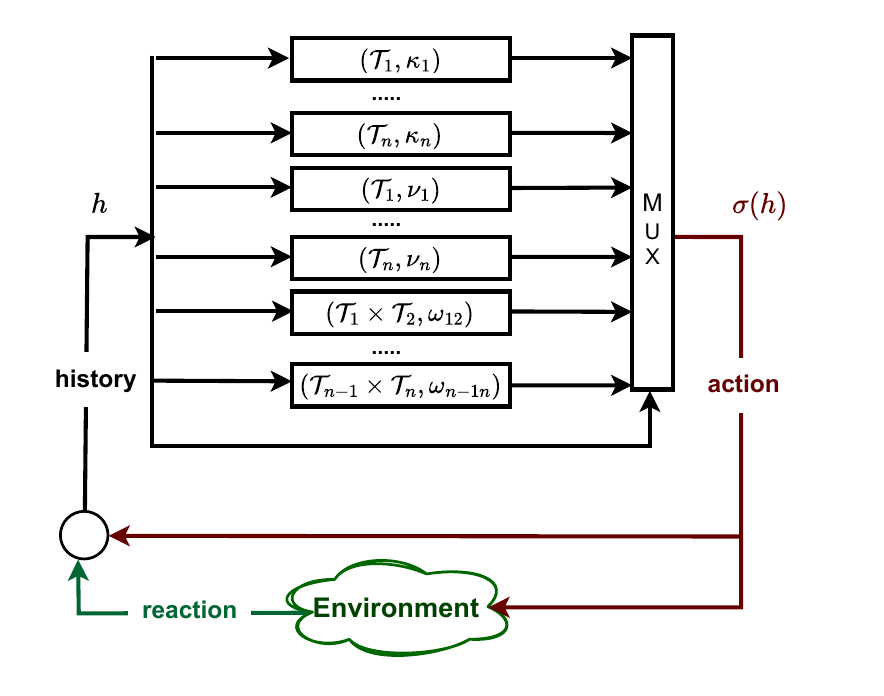}
    \caption{Implementation of the strategy returned by Alg.~\ref{alg:super-mega-iper-ultra-multi-tier}.}
    \label{fig:adaptive-strategy}
\end{figure}

Concretely, the output strategy is implemented by executing simultaneously all transducers~(representing the strategies) computed 
in Line~1.
For the current history $h$, the output strategy $\sigma$ evaluates the response of each such transducer, and selects the output action $\sigma(h)$ based on the tests in Line~2 and discussed above. 
The action $\sigma(h)$, together with the environment reaction, extends $h$, thus generating a new history, and the process starts again.
The strategy terminates when it selects $\bot$, i.e., it is undefined, because either no maximally pending objective exists or the selected transducer outputs $\bot$. 
Figure~\ref{fig:adaptive-strategy} sketches the implementation of the strategy computed with Algorithm~\ref{alg:super-mega-iper-ultra-multi-tier}. 

\begin{theorem}~\label{thm:multi-tier-correct}
    Alg.~\ref{alg:super-mega-iper-ultra-multi-tier} 
    returns an adaptive strategy for $\Phi$ in $\P$.
\end{theorem}


Regarding complexity, Algorithm~\ref{alg:super-mega-iper-ultra-multi-tier} solves games with size \twoexptime and \exptime in that of objectives and domain, respectively.
In fact, the number of solved games is \emph{quadratic} in the number of objectives in the multi-tier goal. 
The following theorem shows the complexity characterization of adaptive synthesis for \LTLf multi-tier goals in nondeterministic planning domains.

\begin{theorem}~\label{thm:complexity}
    \LTLf adaptive synthesis for multi-tier goals is: \begin{compactitem}
        \item \twoexptime-complete in the size of the objectives; 
        \item \exptime-complete in the size of the domain;
        \item \emph{polynomial} in $n$, the number of objectives.
    \end{compactitem}
\end{theorem}

We also note that Line~1 of Algorithm~\ref{alg:super-mega-iper-ultra-multi-tier} is fully parallelizable. 
Synthesizing strategies for each objective can be done in parallel with $n$ processors; synthesizing strategies for pairs of objectives can be done in parallel with $n^2$ processors. As a result, if $n + n^2$ processors are available, adaptive synthesis for multi-tier goals is \emph{virtually} for free, i.e., it costs as handling the most expensive objectives. 
These nice computational results confirm that adaptive synthesis for multi-tier goals brings a minimal overhead to standard synthesis.

\section{Conclusion and Future Work~\label{sec:conclusion}}

In this paper, we introduced the problem of \LTLf adaptive for multi-tier goals in nondeterministic planning domains.
We developed a synthesis technique that is both sound and complete for computing an adaptive strategy in this context. This technique allows for straightforward utilization of symbolic synthesis techniques~\cite{ZTLPV17,DPZ23Ecai}, aiming for promising performance and scalability. Currently, our framework considers a single environment model. However, it is also of interest to consider the setting that involves multiple environment models~\cite{ADLMR20,CiolekDPS20,ADLMR21}, accounting for varying nondeterminism of the environment. For instance, as the environment becomes more nondeterministic, we might anticipate achieving a less challenging objective. We believe that the general approach presented here can be extended to handle this setting as well. 

%

\section*{Acknowledgments}

This work is supported in part by the ERC Advanced Grant WhiteMech (No. 834228), the PRIN project RIPER (No. 20203FFYLK), the PNRR MUR project FAIR (No. PE0000013), and the UKRI Erlangen AI Hub on Mathematical and Computational Foundations of AI. Gianmarco Parretti is supported by the Italian
National Ph.D. on Artificial Intelligence at ``La Sapienza''

\bibliography{aaai25}

\end{document}